# Key Phrase Classification in Complex Assignments

Manikandan Ravikiran

mravikiran3@gatech.edu

*Abstract*—Complex assignments typically consist of open-ended questions with large and diverse content in the context of both classroom and online graduate programs. With the sheer scale of these programs comes a variety of problems in peer and expert feedback, including rogue reviews. As such with the hope of identifying important contents needed for the review, in this work we present a very first work on keyphrase classification with a detailed empirical study on traditional and most recent language modeling approaches. From this study, we find that the task of classification of keyphrases is ambiguous at a human level producing Cohen's kappa of 0.77 on a new dataset. Both pretrained language models and simple TFIDF-SVM classifiers produce similar results with a former producing average of 0.6 F1 higher than the latter. We finally derive practical advice from our extensive empirical and model interpretability results for those interested in key phrase classification from educational reports in the future.

## 1 INTRODUCTION

Peer Feedback and Expert Feedback are inherent parts of Graduate Programs with MOOC form of delivery (Joyner, 2017). While on one hand expert feedback provides valuable and reliable assessment, on the other side peer feedback presents inherent pedagogical benefits. Typically, many of these programs are done at scale with more than hundreds of students enrolled per class, resulting in two side effects first of which is the assignment of multiple students to a single expert owing to high monetary costs to students (Joyner, 2017). On the other hand, peer feedback suffers from a wide variety of issues notably reviews that are insufficient (Geigle, Zhai, and Ferguson, 2016), unusable by peers either because of dishonesty, retaliation, competition or laziness (Kulkarni, Bernstein, and Klemmer, 2015).

Automatic Peer Assessment and Automatic Grading Systems are key in addressing the previously mentioned problems across both Peer Feedback and Expert Feedback, where these systems are restricted to addressing one or more of these problems across multiple stages of feedback. More specifically, we have an automatic scoring system, which directly scores large textual essays or short answers, grading accuracy improvement tools that either adjust the scores based on certain aggregation (Reily, Finnerty, and Terveen, 2009), modeling, calibration and ranking strategies. Then some tools consider better peer and expert allotment (Ardaiz-Villanueva et al., 2011) based on various criteria to get good feedback using grouping strategies. Finally, there are review comment analysis systems that focus on understanding the content of comments to suggest improvements.

While many of these works show positive outcomes with improvements on respective problems (Xiong, Litman, and Schunn, 2010), overall there are still some remaining systematic problems (Geigle, Zhai, and Ferguson, 2016). These include i) changes in reviewers rating with time and length of assignments leading to lack of effective response by the expert ii) discrepancy between rating expected by the author to that of expert and the peer iii) random fluctuations of scores etc. iv) lack of descriptive reviews owing to scale in both peer and expert feedback. These previously mentioned problems worsen further in the context of complex assignments - Assignment with open-ended questions commonly seen in MOOC based graduate programs with large and diverse content.

As such we hypothesize that explicitly identifying the written contents is useful for both peer and expert feedback, thereby could solve the above issues. For example, in the case of complex long assignments, extracting contents that are needed for rubric alone would reduce time spent on the assignment and in turn lead to fruitful reviews. Hence in this work, we propose to classify important phrases of complex assignments?. Since there are plenty of approaches for the classification of educational texts, with varying coverage of accuracy and training benefits. We propose to empirically evaluate the following open research questions.

- **RQ 1.1:** How well do traditional linguistic feature extraction approaches perform on keyphrase extraction? What are the most predictive linguistic features?
- **RQ 1.2:** How well does language models fair in terms of classification of keyphrase from complex assignments?
- **RQ 1.3:** Are the language models more reliable than the traditional feature-based models in the context of keyphrase classification from complex assignments?

In the due process, we introduce a theory grounded annotation scheme, novel dataset, and present exhaustive experimental results. Additionally In line with prior analyses of text classification problems in educational technology we carry out corpus analysis and introduce briefly the methods used for experimentation. Unlike prior efforts, however, our main objectives are to uncover the impact of content and context diversity on performance (measured primarily by F1 score), and also to study the benefits of pre-trained language models over traditional approaches both from the perspective of accuracy and reliability (quantified through accuracy score of interpretability). The rest of the paper is organized as follows, in section 2 we present literature on classification problems and complex assignments, following by datasets and methods used in section 3. Section 4 presents experiments and results. Finally in section 5 we conclude overall findings with some possible implications on future work.

## 2 RELATED WORK

### 2.1 Text Classification & Complex Assignments

Text classification is a long-standing problem in educational technology, which has gone rapid increase with the advent of large scale MOOCs. The works vary according to their final intended goal itself, resulting in a diverse range of datasets, features, and algorithms. The earliest works use a classification approach on click-stream datasets (Yang et al., 2015). Then there is works include that of Scott et al. (2015) which *analyzed three tools*. Then there are also sentiment related works, by (Ramesh et al., 2014) which included linguistic and behavioral features of MOOC discussion forums. Works on similar lines include (Liu et al., 2016), (Tucker, Dickens, and Divinsky, 2014). On the parallel side, some works use posts and their metadata to detect confusion in the educational contents. Notable work by (Akshay et al., 2015), *emphasizes the capacity of posts to improve content creation*. Additionally, there are works on post urgency classification (Omaima, Aditya, and Huzefa, 2018), speech act prediction (Jaime and Kyle, 2015). Finally, some works focus on using classification towards peer feedback, much of which focuses on scaffolding the review comments themselves (Xiong, Litman, and Schunn, 2010), (Nguyen, Xiong, and Litman, 2016), (Ramachandran, Gehringer, and Yadav, 2016), (Cho, 2008).

Similar to above works, we focus on the classification of educational text, however, there is two major difference in our work. First of all being, we focus on assignment text, rather than posts which is predominant. Second, being the cri-teria and end application of classification is peer feedback. Thirdly, rather than scaffolding peer feedback, we focus on extracting textual content suitable for easy peer feedback.

## 2.2 Peer Feedback & Grading

Peer Feedback and grading are integral to online education. There are a plethora of works in this line tackling a wide array of problems. Firstly some works focus on improving grading score accuracy. Notable works, in this line, include that of (Reily, Finnerty, and Terveen, 2009) the work focuses on improving the score through aggregation strategies. Following this, there are also (Piech et al., 2013) which formulated and evaluated a probabilistic peer grading graphical model (PGM) for estimation of submission grades as well as grader bias and reliability. Multiple other works use PGM's models (He, Hu, and Sun, 2019) which use Markov Chain Monte Carlo approaches and (Wu et al., 2015) which uses a Fuzzy Cognitive Diagnosis Framework (Fuzzy CDF).

Alternatively, rather than improving the accuracy of scores, there is also work on the assignment of correct reviewers by forming effective groups of peers. Some of the works in this line (Ardaiz-Villanueva et al., 2011), (Ounnas, 2010). Similarly, there are (Pollalis and Mavrommatis, 2008) which proposes a method for course construction such that collaboration is easily achievable with a locus on common educational goals. Then we have (Lynda et al., 2017) which proposes to address peer assessment based on a combination of profile-based clustering and peer grading with the treatment of results. There are many other works in a similar line with the usage of automatic algorithms to form groups (Graf and Bekele, 2006), (Fahmi and Nurjanah, 2018), and (Bekele, 2006).

In general much of the work focuses on peer grading either towards accuracy improvement, automatic grading, and reviews. However, in our work we focus towards providing evidence-based grading, by classifying important parts of the contents so that peer review is simplified and also evidence is provided along with the scores.

## 3 DATASET, ANNOTATION AND METHODS

In this section, the annotation scheme used for creating the dataset is presented along with algorithms that are used for the experiments.

### 3.1 Annotation Scheme

According to the cognitive theory of writing process (Flower and Hayes, 1981), humans exhibit characteristics of hierarchical writing with the multilevel embedded organization of content where the process is distinctive focusing on goals with continual change, where the writing changes and improves as it progresses. While the original works (Flower and Hayes, 1981), was developed based on general writing, we hypothesize such an observation should be true for complex assignment as well. Further based according to (Hattie and Timperley, 2007) feedbacks are expected to answer major questions that are required from student perspectives including *Where am I going? (What are the goals?), How am I going? (What progress is being made toward the goal?), and Where to next? (What activities need to be undertaken to make better progress?)* (Hattie and Timperley, 2007)

Hence, we use these characteristics (Flower and Hayes, 1981) & (Hattie and Timperley, 2007) to decompose large complex assignments into a series of short important phrases needed for peer review. As such we propose to use the following annotation scheme in Table 1 which consists of four different categories of important phrases that could be extracted from complex assignments. The four different types of phrases include

- **Task** - Representing activity done by the author.
- **Findings** - Indicating the output of activity.
- **Reasons** - Showing reasons behind the findings.
- **Intuition** - Showing background on why the task was executed.

*Table 1*—Annotation Schema for Identification of Important Phrases from Complex Assignment

| Class | Description | Example |
|-------|-------------|---------|
| Task | Task indicates an activity explored by the student | We use using pixel based visual representations for images and develop an production system with series of rules |
| Findings | Findings indicates results of a Task | The agent only solved 2CP's in both sets, which adheres to existing rules, suggesting better rules and analysis are needed to handle CP's |
| Reason | Reason indicates the rationale behind the finding | Approximate similarity property can be seen in BP-E 9, but the result was erroneous |
| Intuition | Intuition represents the reason behind the Task | These problems satisfy simple relationships such as XOR, Overlay, Identity etc. |

Table 2 shows relationship the dataset, annotation scheme, rubrics and works of (Hattie and Timperley, 2007) and (Flower and Hayes, 1981) using KBAI course as example.

| Coding Scheme | Relationship to Rubrics of KBAI | Relationship to (Hattie and Timperley, 2007) | Relationship to (Flower and Hayes, 1981) |
|---|---|---|---|
| *Task* | Project Overview | What are the goals? | Hierarchical, Goal Oriented |
| *Findings* | Cognitive Connection (KBAI) Agent Limitations | Where am I going? | Goal Oriented |
| *Reason* | Relation to KBAI Class | How am I going? | Goal Oriented |
| *Intuition* | Agent Reasoning | What activities need to be undertaken to make better progress? | Hierarchical, Goal Oriented, Distinctive thinking |

## 3.2 Dataset

The dataset in this work was developed using the corpus of the KBAI report of Fall 2019. The overall dataset statistics are as shown in Table 3 below. The data consists of 791 phrases divided into train and test sets respectively. Further considering the comprehensive evaluation of the model. The data is split into two folds used for overall cross-validation. Further, each of the phrases was subject to two rounds of annotation resulting in **Cohen's Kappa($\kappa$) of 0.77** showing that the resultant task is hard and may require more complex semantically relevant features. Further, we can see that the dataset is imbalanced across the four classes with *Task* and *Finding* categories dominating the corpus and *Reason* being the least seen sentences. This behavior is because of two reasons firstly the nature of writing is semi-formal with much of the is on what found as part of results. Secondly, the tokenizer used was from the spacy English model for tokenizing the complex report, occasionally has tokenization errors.

*Table 3*—Dataset statistics and annotation statistics created for this work.

| Splits | Train | | | | Test | | | | Rest |
|---|---|---|---|---|---|---|---|---|---|
| | **T** | **F** | **R** | **I** | **T** | **F** | **R** | **I** | |
| **Fold-1** | 69 | 140 | 25 | 32 | 58 | 90 | 12 | 24 | |
| **Fold-2** | 71 | 140 | 22 | 43 | 56 | 90 | 23 | 15 | |
| **Total** | 276 | | | | 184 | | | | 331 |

Further, the dataset consists of 2443 unique words with occurrences with minimal occurrence of 1 and a maximum occurrence of 800. The combined word statistics are as shown in Figure 1 and words that occur the least amount of time are shown in Figure 2.

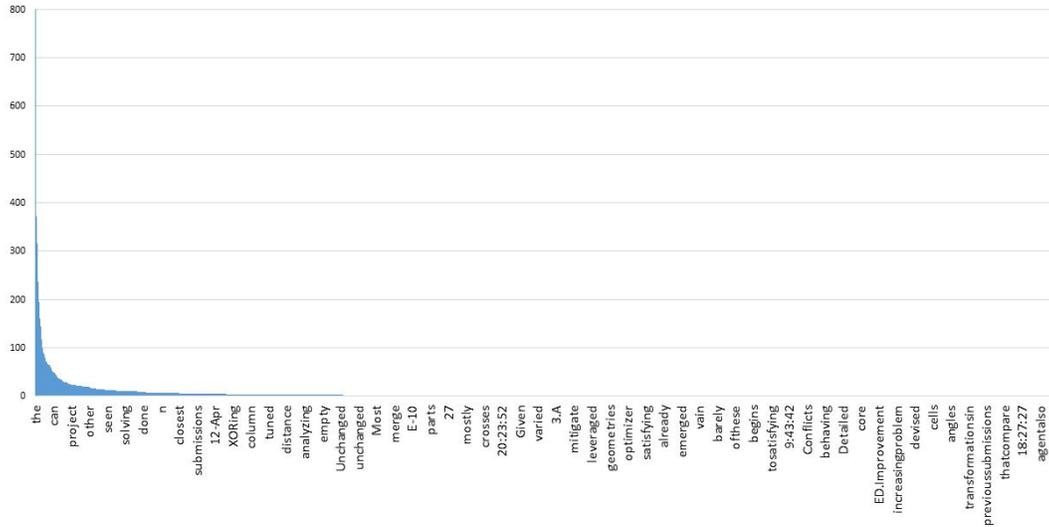

*Figure 1*—Overall word statistics for the developed dataset

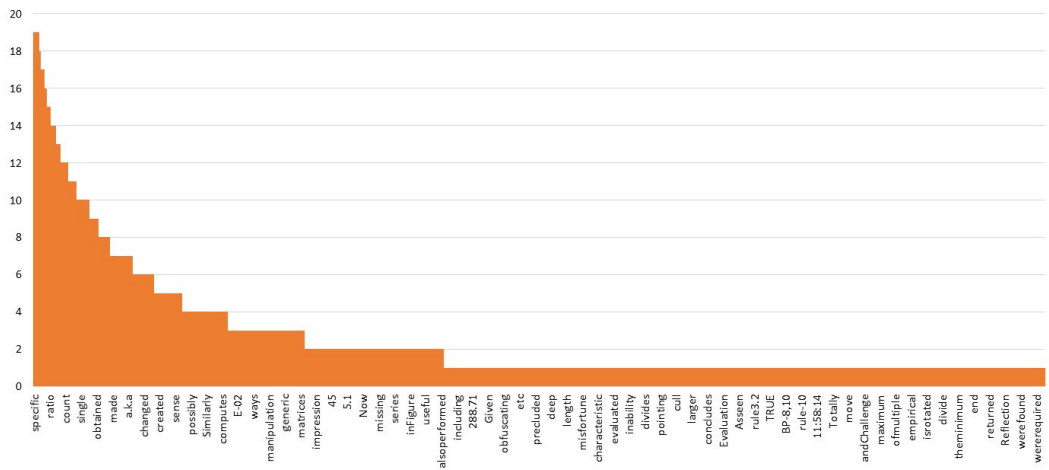

*Figure 2*—Word statistics for words with <20 times of occurrences.

### 3.3 Metrics

We use the following evaluation metrics for phrase classification in this work.

- **Precision (P), Recall (R) and F1-Score (F1):** Precision is the ratio of correctly predicted positive observations to the total predicted positive observations. Recall is the ratio of correctly predicted positive observations to all observations in the actual class. F1 Score is the harmonic mean of Precision and Recall.

### 3.4 Algorithms

**Bert Transformer**: BERT (Devlin et al., 2019) is a trained Transformer Encoder stack, with 12 encoders as part of the base models and 24 in the large version. BERT encoders typically have larger feedforward networks (768 and 1024 nodes in base and large respectively) and more attention heads (12 and 16 respectively). BERT was originally trained on Wikipedia and Book Corpus, a dataset containing +10,000 books of different genres. BERT works like transformer encoder stack, by taking a sequence of words as input which keeps flowing up the stack from one encoder to the next, while new sequences are coming in. The final output for each sequence is a vector of size 728. We will use such vectors for our phrase classification problem without fine-tuning.

**RoBERTA Transformer**: RoBERTA (**Liu2019RoBERTAAR, Liu2019RoBERTAAR**) from Facebook, a robustly optimized BERT approach (RoBERTA), is a retraining of BERT with improved training methodology, 1000% more data and compute power. To improve the training procedure, RoBERTA removes the Next Sentence Prediction (NSP) task from BERT's pre-training and introduces dynamic masking so that the masked token changes during the training epochs. Additionally larger batch size with 160 GB of text for pre-training, including 16GB of Books Corpus and English Wikipedia used in BERT was found to be useful for training. The additional data included Common Crawl News dataset (76 GB), Web text corpus (38 GB) and Stories from Common Crawl (31 GB). This coupled with 1024 V100 Tesla GPU's running for a day, led to the pre-training of RoBERTA. As a result, RoBERTA outperforms both BERT and XLNet on GLUE benchmark results. We again will use the model as the feature extractor.

**Term Frequency:** Term Frequency a.k.a BOW or Term Count measures how frequently a term occurs in a document. Since every document is different in length, multiple tokens would appear much more in long documents than shorter ones. Thus, the term frequency is often divided by the document length (aka. the total number of terms in the document) as a way of normalization.

$$\mathtt{TF(t)} = \frac{\text{Number of times term t appears in a document}}{\text{Total number of terms in the document}} \tag{1}$$

**Term Frequency Inverse document frequency:** The TFIDF algorithm builds on the following representation of sentences/documents. Each sentence d is repre-sented as a vector $d = (d_1, d_2, ...., d_F)$ so that documents with similar content have similar vectors (according to a fixed similarity metric). Each element $d_i$ represents a distinct word $w_i$. $d_i$ for a sentence d is calculated as a combination of the statistics $TF(w_i; d)$ and $DF(w_i)$. The term frequency $TF(w_i; d)$ is the number of times word $w_i$ occurs in document d and the document frequency $DF(w_i)$ is the number of documents in which word $w_i$ occurs at least once. The inverse document frequency $IDF(w_i)$ can be calculated from the document frequency.

$$IDF(w_i) = \log\left(\frac{D}{DF(w_{(i)})}\right) \tag{2}$$

Here, D is the total number of sentences. Intuitively, the inverse document frequency of a word is low if it occurs in many documents and is highest if the word occurs in only one. The so-called weight d(i) of word $w_i$ in sentence $d^i$ is then given as

$$d^{(i)} = TF(w_i; d) IDF(w_i) \tag{3}$$

This word weighting heuristic says that a word $w_i$ is an important indexing term for document d if it occurs frequently in it (the term frequency is high). On the other hand, words that occur in many documents are rated less important indexing terms due to their low inverse document frequency. In this work, we use TF-IDF with Bi-Grams of words.

## 4 EXPERIMENTS

In this section, we present each of the research questions are presented in brief along with their results.

### 4.1 RQ 1.1: Phrase Classification with Traditional Approaches

Our first research question is how traditional classification approaches (SVM) performance differs for the phrases from complex assignment with traditional approaches. To answer this, Precision (P), Recall (R) and F1 metrics are reported on both as weighed average level and original macro-level (See Table 4). The reason for the weighted average is to make results comparable across methods, especially with SVM being sensitive to data imbalance compared to that of gradient boosting models.

For weighted average prediction, the dataset predictions are weighed based on the ratio of the sample of a given category in the dataset. Also as mentioned earlier, to ensure generalization the dataset splits are made such that words in train and test set are diverse and overlapping. Moreover, we grid search all the hyper parameters for both SVM and Gradient boosting algorithms and the TFIDF mechanism with N-grams ranging from 1-3. First look on the dataset suggest the classes of a task, finding and intuition typically encompass words in pairs (predominantly verbs) that could help a given classification approach. Hence we also do search for the N-gram parameter as part of the grid search.

Our best results mentioned in the table are obtained using bi-grams. Further, we implement SVM using gradient descent classifier with Hinge loss, rather than native SVM formation to reduce over fitting and randomness in initialization. This further ensures the overall method doesn't overfit with fewer data. To ensure fair comparison we also create a baseline classifier that predicts class labels proportional to their distribution in the training set. The results of the baseline classifier are again shown in Table 4. We select SVM based on its usage in benchmarks of short text classification (Zeng et al., 2018) and gradient boosting is used due to effectiveness on imbalanced dataset (Wainer and Franceschinell, 2018).

Table 4 shows results with size normalized precision (P), recall (R), and F1- Score (F1). The lowest P, R and F1 values per method are in bold and red to highlight underperformers. Results for all the folds are presented with an accuracy score. Comparing the different methods, the highest F1 results among traditional methods was achieved with SVM with TFIDF, followed by SVM with BOW and Baseline. SVM with TF-IDF shows high precision in Fold-1 and has a balanced P and R in Fold-2, which can be explained by the nature of folds-1, predominantly through our results we observe that the fold-2 is simpler due to similarity in vocabulary compared to that of Fold-1. For SVM with BOW, the results are very close to that of TF-IDF with a small difference of average of 0.2 in the F1 score. These findings are in line with other work reporting the usefulness of SVM in short text classification  (Zeng et al., 2018).

Concerning results on each of the folds, SVM with BOW produces similar results but again dominated by SVM with TF-IDF by a large margin (>= 6 points in F1) The first reason success can be attributed to being from the same the dataset, hence similar vocabulary. However, at the same time, the performance is lower compared to that of typical text classification mostly because of the semi-formal nature of the text. This could be linked to noise typically seen in social media texts.

*Table 4*—Overall results for phrase classification for RQ 2.1-2.2

| RQs | Approaches | Fold -1 | | | | | | | Fold -2 | | | | | | |
| | | | Macro Average | | | Weighed Average | | | | Macro Average | | | Weighed Average | | |
| | | A | P | R | F1 | P | R | F1 | A | P | R | F1 | P | R | F1 |
| | Baseline (Distribution Based Classifier) | 0.4 | 0.29 | 0.3 | 0.29 | 0.39 | 0.4 | 0.38 | 0.34 | 0.25 | 0.24 | 0.24 | 0.36 | 0.34 | 0.35 |
| RQ 2.1 | SVM + BOW | 0.54 | 0.46 | 0.41 | 0.42 | 0.53 | 0.54 | 0.52 | 0.54 | 0.38 | 0.37 | 0.37 | 0.51 | 0.54 | 0.52 |
| | SVM + TFIDF | 0.61 | 0.57 | 0.4 | 0.42 | 0.61 | 0.61 | 0.56 | 0.592 | 0.45 | 0.41 | 0.42 | 0.64 | 0.59 | 0.61 |
| RQ 2.2 | BERT (SVM) | 0.56 | 0.47 | 0.4 | 0.45 | 0.57 | 0.56 | 0.52 | 0.6 | 0.51 | 0.5 | 0.48 | 0.63 | 0.6 | 0.61 |
| | BERT (XGB) | 0.56 | 0.49 | 0.37 | 0.38 | 0.54 | 0.56 | 0.52 | 0.56 | 0.49 | 0.37 | 0.38 | 0.54 | 0.56 | 0.52 |
| | RoBERTa (SVM) | 0.13 | 0.03 | 0.25 | 0.05 | 0.01 | 0.13 | 0.03 | 0.51 | 0.39 | 0.31 | 0.28 | 0.53 | 0.51 | 0.42 |
| | RoBERTa (XGB) | 0.57 | 0.49 | 0.36 | 0.37 | 0.57 | 0.57 | 0.52 | 0.55 | 0.36 | 0.34 | 0.33 | 0.5 | 0.55 | 0.49 |

Our overall analysis of traditional method shows that it performs fairly well on the complex assignment phrases, however owing to sparsity in vocabulary and size of dataset the performance is not so high like usual text classification approaches. Hence, with higher datasets and different domains, the performance is expected to be consistent. To summarise, our findings are:

• F1 is highest with SVM with TF-IDF, followed by SVM with BOW.
• SVM with TF-IDF outperforms other traditional methods so compared by a large margin (e.g. >= 6 points in F1)
• Existing hypothesis of SVM with TF-IDF serves as a competitive baseline for short text classification still holds.

## 4.2 RQ 1.2: Phrase Classification with Language Models

While SVM with TF-IDF does produce significant results, but the error and ambiguity are extremely. This is visible from results F1-Score which is below 0.5 F1-Score. Recently, language models offer significant advantage where they help to build on existing knowledge gathered from large datasets since much of the dataset is derived from semi-formal reports we hypothesize such pretrained networks would help in results. Further, since the dataset is very small, training will lead to significant over fitting hence to alleviate this we propose to use the pretrained language models only as feature extractors with SVM and gradient boosting classifier in line with RQ 1.1

Table 4 shows the results on the language models with both BERT and RoBERTA transformer base models used as feature extractors. The features are extractor from the second encoder of the transformer (this was based on our experimentation with different encoders). Firstly comparing BERT with SVM and RoBERTA with SVM we can see that BERT offers a top performance of 0.48 F1-Score and RoBERTA produces max results of 0.37 F1-Scores. Secondly, we can see that RoBERTA offers an extremely bad result in Fold-1 while the results are higher in fold-2. Previously in RQ 1.1, we mentioned that Fold-1 is harder than fold-2, which is the main reason. This result was tested multiple times and no issue or correlation level issues were found on Fold-1 with RoBERTA. To further check the impact of training, we froze all the encoders and trained only the final layers, yet the results were significantly lower than the baseline. Additionally, we can see from Table Table 4 results using XGB, the performance is pretty much similar across the cases of BERT and RoBERTA, where unlike SVM the results are balanced and high.

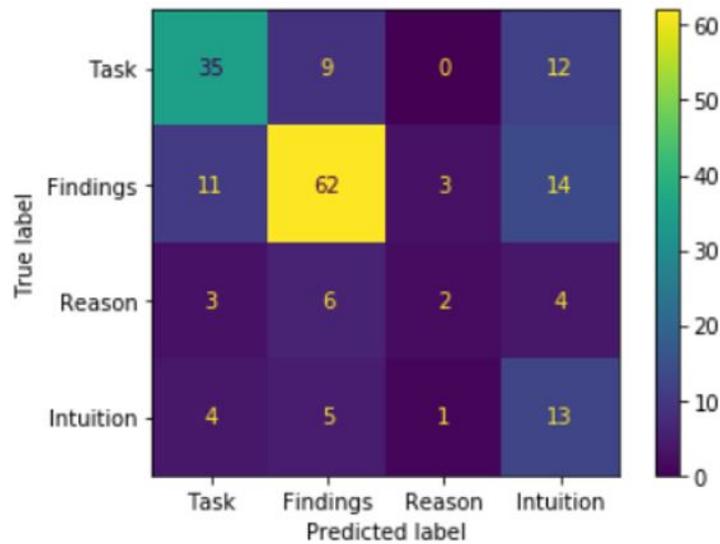

*Figure* 3—Confusion Matrix of the best performing BERT-SVM model

Regarding RoBERTA's performance on SVM in Fold0-1 we suspect following, RoBERTA originally uses a special training procedure with dataset larger than that of BERT, while we expect the pretrained representations to be well distributed, however with training on large sets the representations tend to form tight clusters results in need of greater weighting of data points. However such details needs to be validated. At the same, time the experimental setup needs to be revisited especially the RoBERTA tokenizer.

Finally, coming to errors we can see that the classes of *Task, Finding, and Intuition* are highly ambiguous. Figure 3 shows the confusion matrix, where we can see much of the error is concentrated among the three classes. This is in line with Cohen's kappa (κ) of 0.77 earlier mentioned in the dataset.

To summarise, our findings are:

- F1 on RoBERTA is significantly lower than F1 on BERT for all 2 folds, which is mostly(speculation) due to inherent architectural flaw. However experimental setup needs to be revisited.
- Performance on BERT is significantly and consistently higher than that of RoBERTA and best-performing SVM in different folds, with the lower scores mostly attributable to lower recall. Further, we can see that XGB with either representation has no significant changes in terms of results. A reason that needs to be revisited.
- However, there are still significant differences in the classification of phrases and human annotation. Considering Cohen's kappa (κ) of 0.77 we are far from results. Capturing context is a hard issue especially with phrases of complex assignments and current annotation schemes. Thus the annotation and model architectures need to be revisited.

### 4.3  RQ 1.3: Reliability study of traditional and language models

Previously in Table 4, we saw the results of both traditional and language models. We could see that among traditional models the performance was closer with TF-IDF producing higher performance on fold-2. Then coming to language models, they tend to perform significantly higher than the rest. So to further evaluate the performance in-depth, we focused on the reliability aspect of the model where we define *reliability is defined as the models ability to focus on the required information for classification that is true from a human's perspective* i.e. the model should look on some parts of the text which a human would see during annotation. To do this, we employ Local Interpretable Model-agnostic Explanations(LIME) (Ribeiro, Singh, and Guestrin, 2016). For this, we modify the original implementation of the language models for a single forward pass, by decoupling the pre-processing provided by the Hugging Face Transformer Library.

We use the **precision** as the evaluation metric, where we compute precision as the fraction of words that are contributing to both traditional and language models performance. The results so obtained are as shown in Table 5. Further, we manually analyze the results to find interesting characteristics. Figure 4 shows results where both traditional and language models look on similar words, while Figure 5 shows the BERT model to focus on words that don't make any sense in terms of contributions. We don't use XGB and RoBERTA models as they tend to produce lower performance.

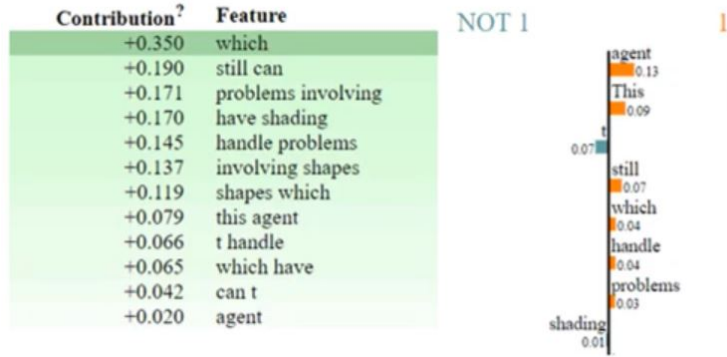

*Figure 4*—Example of Interpretability of Traditional (right) and Language models (left) with similar word contributions.

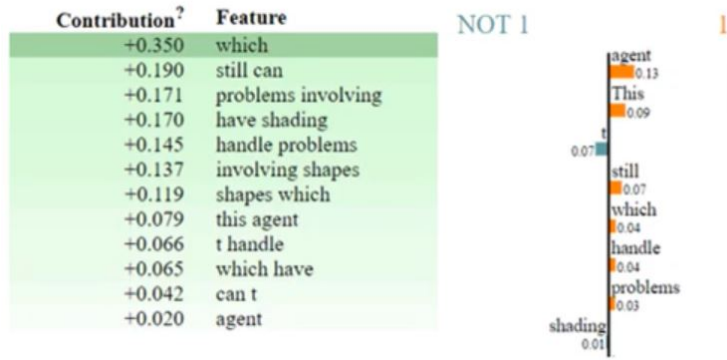

*Figure 5*—Example of Interpretability of Traditional (right) and Language models (left) with dissimilar word contributions.

*Table 5*—Results of Interpretability on Traditional and BERT

|   | Model -1 | Model-2 | Precision |
|---|----------|---------|-----------|
| 1 | BERT (SVM) | SVM (TF-IDF) | 0.82 |
| 2 | BERT (SVM) | SVM (BOW) | 0.81 |

From a results point of view, We find that in both the cases the models look around 80% of times on similar word lists. Further, we also see mixed interpretability The results of interpretability are inconclusive were in. From Figure 4 we can see that both traditional and BERT look for similar words for predicting the correct results, however, from Figure 5 we can see that in some cases traditional model has better interpretability. We speculate this is an issue of using the BERT model as a feature extractor, rather than training it from scratch. Overall traditional models tend to focus on the right set of words, in the majority of cases. Hence we conclude that traditional models better interpret the cues related to class from the educational text. However, at the same time, we argue that this needs to be revisited in more depth with an exhaustive comparison on a relationship with features used in the traditional model and language model training process.

To summarise, our findings are as follows:

- Both traditional and language models look on similar word cues around 80% of times, despite dissimilar contributions.
- Traditional models are more reliable in terms of interpretability, this has more to do with the usage of simple features itself. Meanwhile, in language models, we use complex encoders without any training.

## 5 CONCLUSION

This paper investigated the ability of both traditional and recent language models for the task of classification of phrases from complex assignments. Firstly, by developing a new corpus and annotation scheme, we demonstrated that datasets in complex assignments are harder to create: in terms of annotation and size; the balance of classes; the proportion of words; and how often tokens are repeated. The dataset so created is imbalanced with the domination of *Task* and *Finding* classes. This is similar, to the dataset of twitter and web corpora which are traditionally noisy. More interestingly, however, despite imbalance, the task is inherently harder as the cues are very limited with lower information density as the classification is fine-grained.

Our second set of findings relates to the traditional approaches studied. In this work, we studied only SVM with BOW and TF-IDF approaches. We find that SVM with TF-IDF achieves consistently the highest performance across the folds, and this is the best traditional approach to generalizing from training to testing data. This can mostly be attributed to SVM"s inherent quality when trained with word-level features. Further, the tuned parameters of SVM achieve a balanced precision and recall.

Thirdly, what is found to be a good predictor of F1 is a BERT trained with SVM, in which the BERT model was used as a feature extractor for each fold of the test corpus. This supported our hypothesis that language models without any training would still be useful owing to their training on large datasets, and training on a small dataset will negatively be correlated with F1. While we didn't study details on proportions of unseen token across train and test dataset, we can argue that the majority of terms across the classes that are important include simple ones such as verbs and verb phrases.

Fourth, our experiments also confirmed the issue of reliability where traditional models are more reliable in terms of interpretability while language models without training are sensitive in nature. In particular, we saw that the net precision of common words used as cues is 0.82 which indicates that the task could be addressed well by simple algorithms at the same time it requires complex features, rather than simple TF-IDF.

Finally, by studying performance with the weighted average, it becomes clear that there is also a big difference in performance on corpora with the balanced and imbalanced dataset. This indicates that annotating more training examples for diverse classes would likely lead to a dramatic increase in F1 which in turn is expected to improve performance across all the explored RQ's.

## 6 REFERENCES

[1]    Akshay, A., Jagadish, V., Shane, L., and Andreas, P. (2015). "YouEDU: Addressing Confusion in MOOC Discussion Forums by Recommending Instructional Video Clips". In: *Proceedings of the Eighth International Conference on Educational Data Mining (EDM 2015)*. Madrid, Spain.

[2]    Ardaiz-Villanueva, Oscar, Chacón, Xabier Nicuesa, Artazcoz, Oscar Brene, Acedo Lizarraga, María Luisa Sanz de, and Acedo Baquedano, María Teresa Sanz de (2011). "Evaluation of computer tools for idea generation and team formation in project-based learning". In: *Computers  Education* 56, pp. 700–711.

[3] Bekele, Rahel (2006). "Computer-Assisted Learner Group Formation Based on Personality Traits". In: *Dissertationsschrift zur Erlangung des Grades eines Doktors der Naturwissenschaften am Fachbereich Informatik der Universität Hamburg*.

[4] Cho, Kwangsu (2008). "Machine Classification of Peer Comments in Physics". In: *First International Conference on Educational Data Mining*.

[5] Devlin, J., Chang, M., Lee, K., and Toutanova, K. (2019). "BERT: Pre-training of Deep Bidirectional Transformers for Language Understanding". In: *Proceedings of the 2019 Conference of the North American Chapter of the Association for Computational Linguistics: Human Language Technologies*.

[6] Fahmi, Fitra Zul and Nurjanah, Dade (2018). "Group Formation Using Multi Objectives Ant Colony System for Collaborative Learning". In: *2018 5th International Conference on Electrical Engineering, Computer Science and Informatics (EECSI)*, pp. 696–702.

[7] Flower, Linda and Hayes, J. R. (1981). "A Cognitive Process Theory of Writing." In: *College Composition and Communication Vol. 32, No. 4, pp. 365-387*.

[8] Geigle, Chase, Zhai, ChengXiang, and Ferguson, Duncan C. (2016). "An Exploration of Automated Grading of Complex Assignments". In: *Proceedings of the Third International Conference of on Learning at Scale (L@S)*. Edinburgh, United Kingdom.

[9] Graf, Sabine and Bekele, Rahel (2006). "Forming Heterogeneous Groups for Intelligent Collaborative Learning Systems with Ant Colony Optimization". In: *Intelligent Tutoring Systems*.

[10] Hattie, John and Timperley, Helen (2007). "The Power of Feedback". In: *Review of Educational Research* 77.1, pp. 81–112. DOI: 10.3102/003465430298487. eprint: https://doi.org/10.3102/003465430298487. URL: https://doi.org/10.3102/003465430298487.

[11] He, Yu, Hu, Xinying, and Sun, Guangzhong (2019). "A cognitive diagnosis framework based on peer assessment". In: *ACM Turing Celebration Conference*.

[12] Jaime, A. and Kyle, S. (2015). "Predicting Speech Acts in MOOC Forum Posts". In: *International AAAI Conference on Web and Social Media*. Oxford, England. URL: https://www.aaai.org/ocs/index.php/ICWSM/ICWSM15/paper/view/10526.

[13]   Joyner, David A. (2017). "Scaling Expert Feedback: Two Case Studies". In: *Proceedings of the Fourth International Conference of on Learning at Scale (L@S)*. Cambridge, MA.

[14]   Kulkarni, Chinmay, Bernstein, Michael S., and Klemmer, Scott R. (2015). "PeerStudio: Rapid Peer Feedback Emphasizes Revision and Improves Performance". In: *Proceedings of the Second International Conference of on Learning at Scale (L@S)*. Vancouver, British Columbia.

[15]   Liu, Z., Liu, S., Liu, L., Sun, J., Peng, X., and Wang, T. (2016). "Sentiment recognition of online course reviews using multi-swarm optimization-based selected features". In: *Neurocomputing* 185, pp. 11–20.

[16]   Lynda, Haddadi, Farida, Bouarab-Dahmani, Tassadit, Berkane, and Samia, Lazib (2017). "Peer assessment in MOOCs based on learners' profiles clustering". In: *2017 8th International Conference on Information Technology (ICIT)*, pp. 532–536.

[17]   Nguyen, Huy, Xiong, Wenting, and Litman, Diane J. (2016). "Instant Feedback for Increasing the Presence of Solutions in Peer Reviews". In: *Proceedings of the 15th Annual Conference of the North American Chapter of the Association for Computational Linguistics: Human Language Technologies*.

[18]   Omaima, A., Aditya, J., and Huzefa, R. (2018). "Needle in a haystack: Identifying learner posts that require urgent response in MOOC discussion forums". In: *Computers  Education* 118, pp. 1–9.

[19]   Ounnas, Asma (2010). "Enhancing the automation of forming groups for education with semantics". In: *Doctoral Thesis, School of Electronics and Computer Science,University of Southampton*.

[20]   Piech, Chris, Huang, Jonathan, Chen, Zhenghao, Do, Chuong B., Ng, Andrew Y., and Koller, Daphne (2013). "Tuned Models of Peer Assessment in MOOCs". In: *ArXiv* abs/1307.2579.

[21]   Pollalis, Yannis A. and Mavrommatis, George (2008). "Using similarity measures for collaborating groups formation: A model for distance learning environments". In: *European Journal of Operational Research* 193, pp. 626–636.

[22]   Ramachandran, Lakshmi, Gehringer, Edward F., and Yadav, Ravi (2016). "Automated Assessment of the Quality of Peer Reviews using Natural Language Processing Techniques". In: *International Journal of Artificial Intelligence in Education* 27, pp. 534–581.

[23]   Ramesh, Arti, Goldwasser, Dan, Huang, Bert, Daumé, Hal, and Getoor, Lise (2014). "Understanding MOOC Discussion Forums using Seeded LDA". In: *Proceedings of the Ninth Workshop on Innovative Use of NLP for Building Educational Applications*. Baltimore, Maryland.

[24]   Reily, Ken, Finnerty, Pam Ludford, and Terveen, Loren G. (2009). "Two peers are better than one: aggregating peer reviews for computing assignments is surprisingly accurate". In: *ACM SIGCHI International Conference on Supporting Group Work, GROUP'09*.

[25]   Ribeiro, Marco Tulio, Singh, Sameer, and Guestrin, Carlos (2016). ""Why Should I Trust You?": Explaining the Predictions of Any Classifier". In: *KDD '16*.

[26]   Scott, A. C., M., Danielle S., Ryan, S., Yuan, W., Luc, P., Tiffany, B., and Yoav, B. (2015). "Language to Completion: Success in an Educational Data Mining Massive Open Online Class". In: *Eighth International Conference on Educational Data Mining*. Madrid, Spain.

[27]   Tucker, C. S., Dickens, B., and Divinsky, A. (2014). "Knowledge Discovery of Student Sentiments in MOOCs and Their Impact on Course Performance". In: *Proceedings of the ASME Design Engineering Technical Conference*. Buffalo, NY, USA.

[28]   Wainer, Jacques and Franceschinell, Rodrigo A. (2018). "An empirical evaluation of imbalanced data strategies from a practitioner's point of view". In: *ArXiv* abs/1810.07168.

[29]   Wu, Run-ze, Liu, Qi, Liu, Yuping, Chen, Enhong, Su, Yu, Chen, Zhigang, and Hu, Guoping (2015). "Cognitive Modelling for Predicting Examinee Performance". In: *Proceedings of the Twenty-Fourth International Joint Conference on Artificial Intelligence*.

[30]   Xiong, Wenting, Litman, Diane J., and Schunn, Christian D. (2010). "Assessing Reviewer's Performance Based on Mining Problem Localization in Peer-Review Data". In: *Proceedings of 3rd International Conference on Educational Data Mining*.

[31]   Yang, D., Wen, M., Howley, I. K., Kraut, R. E., and Penstein, C. (2015). "Exploring the Effect of Confusion in Discussion Forums of Massive Open Online Courses". In: *Proceedings of the Second International Conference on Learning at Scale (L@S)*. Vancouver, British Columbia, Canada: ACM Press.

[32]   Zeng, Jichuan, Li, Jing, Song, Yan, Gao, Cuiyun, Lyu, Michael R., and King, Irwin (2018). "Topic Memory Networks for Short Text Classification". In: *EMNLP*.